\documentclass[letterpaper, 10 pt, conference]{ieeeconf}  

\IEEEoverridecommandlockouts                              

\usepackage{graphicx}
\usepackage{amsmath} 
\usepackage{cite}
\usepackage{hyperref}
\usepackage{cleveref}
\usepackage{acro}
\usepackage{booktabs}
\usepackage{tabularx}
\usepackage[table]{xcolor} 

\DeclareAcronym{auv}{
  short = AUV,
  long = Autonomous Underwater Vehicle
}
\DeclareAcronym{fps}{
  short = FPS,
  long = Frame per Second
}
\DeclareAcronym{fsm}{
  short = FSM,
  long = Finite State Machine
}
\DeclareAcronym{drl}{
  short = DRL,
  long = Deep Reinforcement Learning
}
\DeclareAcronym{ins}{
  short = INS,
  long = Inertial Navigation System
}
\DeclareAcronym{mdp}{
  short = MDP,
  long = Markov Decision Process
}
\DeclareAcronym{iusbl}{
  short = I-USBL,
  long = Inverted Ultra-short baseline
}

\DeclareAcronym{slam}{
  short = SLAM,
  long = Simultaneous Localization and Mapping
}
\DeclareAcronym{soa}{
  short = SoA,
  long = State of the Art
}
\DeclareAcronym{iauv}{
  short = I-AUV,
  long = Intervention Autonomous Underwater Vehicle
}
\DeclareAcronym{3dbm}{
  short = 3DBM,
  long = Three Dimensional Binary Marker
}
\DeclareAcronym{3dam}{
  short = 3DAM,
  long = Three Dimensional ArUco Marker
}
\DeclareAcronym{rov}{
  short = ROV,
  long = Remotely Operated Vehicle
}
\DeclareAcronym{ds}{
  short = DS,
  long = Docking Station
}
\DeclareAcronym{um}{
  short = UM,
  long = Underwater Marker
}
\DeclareAcronym{ltd}{
  short = LTD,
  long = Long-term Deployment
}
\DeclareAcronym{ahrs}{
  short = AHRS,
  long = Attitude and Heading Reference System
}
\DeclareAcronym{msc}{
  short = MSC,
  long = Managed Surge Controller
}
\DeclareAcronym{usbl}{
  short = USBL,
  long = Ultra Short Baseline
}
\DeclareAcronym{hrp}{
  short = HRP,
  long = Humanoid Robotics Platform
}
\DeclareAcronym{uav}{
  short = UAV,
  long = Unmanned Aerial Vehicle
}
\DeclareAcronym{asv}{
  short = ASV,
  long = Autonomous Surface Vehicle
}
\DeclareAcronym{ml}{
  short = ML,
  long = Machine Learning
}
\DeclareAcronym{mpc}{
  short = MPC,
  long = Model Predictive Control
}
\DeclareAcronym{lbl}{
  short = LBL,
  long = Long Baseline
}
\DeclareAcronym{pid}{
  short = PID,
  long = Proportional-Integral-Derivative Controller
}
\DeclareAcronym{cola2}{
  short = COLA2,
  long = Component-Oriented Layer-based Architecture for Autonomy
}
\DeclareAcronym{ros}{
  short = ROS,
  long = Robot Operating System
}
\DeclareAcronym{imu}{
  short = IMU,
  long = Inertial Measurement Unit
}
\DeclareAcronym{dvl}{
  short = DVL,
  long = Doppler Velocity Log
}
\DeclareAcronym{gps}{
  short = GPS,
  long = Global Positioning System
}
\DeclareAcronym{3d}{
  short = 3D,
  long = Three dimensions
}
\DeclareAcronym{2d}{
  short = 2D,
  long = Two dimensions
}
\DeclareAcronym{ekf}{
  short = EKF,
  long = Extended Kalman Filter
}
\DeclareAcronym{pnp}{
  short = PnP,
  long = Perspective-n-Point Pose Computation
}
\DeclareAcronym{ned}{
  short = NED,
  long = {North, East, and Down}
}
\DeclareAcronym{urdf}{
  short = URDF,
  long = Unified Robotics Description Format
}
\DeclareAcronym{gnc}{
  short = GNC,
  long = ¨{Guidance, Navigation and Control}
}
\DeclareAcronym{gds}{
  short = GDS,
  long = Girona Docking Station
}
\DeclareAcronym{sds}{
  short = SDS,
  long = Sparus Docking Station
}
\DeclareAcronym{cad}{
  short = CAD,
  long = Computer-Aided Design
}
\DeclareAcronym{vicorob}{
  short = ViCOROB,
  long = Computer Vision and Robotics
}
\DeclareAcronym{cirs}{
  short = CIRS,
  long = Centro de Investigación en Robótica Submarina
}
\DeclareAcronym{udg}{
  short = UDG,
  long = Universitat de Girona
}
\DeclareAcronym{petg}{
  short = PETG,
  long = Poly-Ethylene Terephthalate Glycol
}
\DeclareAcronym{obsea}{
  short = OBSEA,
  long = Expandable Seafloor Observatory
}
\DeclareAcronym{dof}{
  short = DoF,
  long = Degree of Freedom
}
\DeclareAcronym{vlc}{
  short = VLC,
  long = Visible Light Communication
}
\DeclareAcronym{wic}{
  short = WIC,
  long = Wireless Inductive Charger
}
\usepackage{amssymb}
\usepackage{subcaption}
\usepackage[normalem]{ulem}

\usepackage{supertabular}
\usepackage{colortbl}

\title{\LARGE \bf
Sim-to-reality adaptation for Deep Reinforcement Learning applied to an underwater docking application}

\author{Alaaeddine Chaarani, Narcis Palomeras, and Pere Ridao$^{1}$
\thanks{$^{1}$Authors with the Faculty of Computer Science, Computer Vision and Robotics Research Group (vicorob)
        Universitat de Girona, 17003 Girona, Spain.
        }%
}

\begin{document}

\maketitle
\thispagestyle{empty}
\pagestyle{empty}

\begin{abstract}

Deep Reinforcement Learning (DRL) offers a robust alternative to traditional control methods for autonomous underwater docking, particularly in adapting to unpredictable environmental conditions. However, bridging the "sim-to-real" gap and managing high training latencies remain significant bottlenecks for practical deployment. This paper presents a systematic approach for autonomous docking using the Girona Autonomous Underwater Vehicle (AUV) by leveraging a high-fidelity digital twin environment. We adapted the Stonefish simulator into a multiprocessing RL framework to significantly accelerate the learning process while incorporating realistic AUV dynamics, collision models, and sensor noise. Using the Proximal Policy Optimization (PPO) algorithm, we developed a 6-DoF control policy trained in a headless environment with randomized starting positions to ensure generalized performance. Our reward structure accounts for distance, orientation, action smoothness, and adaptive collision penalties to facilitate soft docking. Experimental results demonstrate that the agent achieved a success rate of over $90\%$ in simulation. Furthermore, successful validation in a physical test tank confirmed the efficacy of the sim-to-reality adaptation, with the DRL controller exhibiting emergent behaviors such as pitch-based braking and yaw oscillations to assist in mechanical alignment.

\end{abstract}

\section{INTRODUCTION}
Modern applications are increasingly utilizing \ac{ml} to achieve superior results and more generalized behavior in autonomous tasks. While certain actions remain achievable through standard methods such as \ac{pid} or \ac{mpc}, \ac{drl} offers the distinct advantage of adapting to unaccounted-for environmental conditions. This robustness is the primary motivation for deploying \ac{drl} across multiple platforms, including \acp{asv}, \acp{uav}, and \acp{hrp}.

Currently, the primary bottlenecks for \ac{drl} are training latency and the "sim-to-real" gap. Various approaches exist to accelerate training and achieve optimal policies using high-performance platforms like MJX \cite{mjx} and Isaac Sim \cite{mittal2025isaaclab}. However, while training efficiency is rapidly improving, sim-to-real adaptation remains in its early stages. This research focuses on facilitating a seamless transfer from simulation to reality by leveraging high-fidelity environments.

Underwater applications such as manipulation and docking can significantly benefit from \ac{drl}-based control strategies. In this work, we address the problem of autonomous docking using the Girona \ac{auv} \cite{girona}. This task involves several challenges inherent to underwater environments, making it a suitable testbed for evaluating the adaptive capabilities of modern \ac{drl} approaches.

Recent advances in \ac{drl}-based docking aim to achieve robust performance under challenging conditions, such as ocean currents and sensor noise, where traditional control strategies often degrade. Nevertheless, only a limited number of studies have successfully transferred policies trained in simulation to real underwater platforms. 

The main contributions of this paper are as follows:

\begin{itemize}
\item Adapting the Stonefish simulation to a multiprocessing RL framework, significantly accelerating the learning process.
\item Developing a high-fidelity environment within Stonefish that incorporates \ac{auv} dynamics, precise collision models, and realistic sensor inputs to facilitate sim-to-real adaptation.
\item Integrating position-based servoing with \ac{drl} as a robust replacement for standard control systems and behavior trees.
\item Demonstrating successful autonomous docking in a physical test tank using the proposed \ac{drl} methodology.
\end{itemize}

\begin{table*}[ht]
\centering
\caption{Summary of \ac{drl} used for AUV control and Docking, check appendix for acronyms.}
\label{tab:rl_SOA}
\small
\begin{tabularx}{\textwidth}{@{}l X X c X l l@{}}
\toprule
\textbf{Reference} & \textbf{Algorithm} & \textbf{Observation} & \textbf{Action} & \textbf{Reward} & \textbf{Sim.} & \textbf{Real Env.} \\ \midrule

Anderlini (2019) \cite{Anderlini2019} & DDPG, DQN & $[x_r, z_r, \theta_r, \dot{x}, \dot{z}, \dot{\theta}, n]$ & $[Q_m, \delta_s]$ & Distance, Success & Matlab & None \\ \addlinespace

Zhang (2020) \cite{Zhang2020} & DQNH/E & $[d, c, k, c_d]$ & Rudder values  & Custom reward & Gazebo & None \\ \addlinespace

Patil (2021) \cite{Patil2021} & TD3, SAC, PPO & [x, y, $\psi$, u, v, r,
$n_1, n_2, n_3$] & $[n_1,n_2,n_3]$ & $R_{dist}, R_{thrust},$
$ R_{align}$ & UUV Simulator & None \\ \addlinespace

Zhang (2023) \cite{Zhang2023} & ARDR, ABPPO, SAC, TD3, PPO  & $[\Delta d_r, \Delta d_y, \Delta \psi_c, $
$ \Delta \psi_g, \theta, \psi, f, \delta_1, \delta_2]$ & $[f, \delta_1, \delta_2]$ & Reward, penalty & Custom & None \\ \addlinespace

Bharti (2025) \cite{Bharti2025} & TD3 & $[e_{1\times6}, F_{1\times4}]$ & $[F_{xk}, F_{yk}, F_{zk}, \tau_{zk}]$ & $r_d, r_{yaw}, r_{bear}, $
$r_{elev}, r_{smooth}$ & Gazebo & Test tank \\ \addlinespace

Zheng (2025) \cite{Zheng2025} & SAC & $[u_t, v_t, r_t, \chi_t, e_{y,t}, \delta_t]$ & Rudder values & $[e_x,e_y,\delta]$ & Custom  & None \\ \addlinespace

Yu \& Lin (2025) \cite{Yu2025} & TD3 + DDPG + YOLO & $[T_u,T_v,W_b,h_b]$ & Rudder values &  $r_h + e_{\delta H} + r_{\psi} + r_y$ & MATLAB & Test Tank \\ \addlinespace

Chu (2025) \cite{Chu2025} & ARSPPO & $[d_t,\cos(\psi_{r,t}),\sin(\psi_{r,t}),$
$\psi_t, u_t,w_t,r_t]$ & $[f_t,m_t]$ & $r_{dist}+r_{post}+r_{action}+r_{time}$ & Gazebo & Lake \\ \addlinespace

Tuncay (2026) \cite{Tunay2026-2} & SHAC, PPO, DroQ, MPC & $[\delta_d,\delta_\tau,V,W]$ & $[F_x,F_y,F_z,$ 
$\tau_\phi, \tau_\theta, \tau_\psi]$ & $r_{\text{pos}} + r_{\text{att}} + r_{\text{act}} + r_{\text{vel}} + r_{\text{act-mavg}}$ & MJX / Stonefish & Test Tank \\ 
\bottomrule
\end{tabularx}
\end{table*}

\section{State of the art}

The development of \ac{drl} for underwater docking has evolved significantly over the last few years, moving from basic kinematic benchmarks to complex sim-to-real implementations.

Early work by Anderlini et al. (2019) \cite{Anderlini2019} presented one of the first adaptations of RL to the docking problem. They compared DDPG and DQN against traditional PID and optimal control methods. To ensure a fair comparison, they utilized a fixed starting position and a standardized dynamic model based on the RAMMUS 100 \ac{auv}. Shortly after, Zhang et al. (2020) \cite{Zhang2020} explored path following by combining \ac{drl} with interactive RL. They introduced human-in-the-loop reward structures (DQNH and DQNHE) to accelerate the learning process compared to standard DQN, evaluating their results on the horizontal plane.

In 2021, Patil et al. \cite{Patil2021} established a benchmark for continuous docking control using a torpedo-shaped \ac{auv} and a fixed docking station. Their study evaluated PPO, TD3, and SAC, concluding that TD3 provided the most robust performance. The reward structures utilized in their benchmarking are further detailed in \Cref{tab:rl_SOA}.

More recent studies have shifted focus toward environmental robustness and 3D complexity. Zhang et al. (2023) \cite{Zhang2023} utilized PPO to perform docking under variable wave and current conditions. While their 3D environment closely mirrored real-world physics, they noted that the absence of a dedicated collision model meant that any contact resulted in immediate task termination—a factor that can significantly impact docking success rates.

The transition from simulation to physical hardware represents the current frontier of the field. In \cite{Palomeras24}, several practical strategies to facilitate the transition from simulation to real-world deployment were proposed. However, it was not until the work of Bharti et al. \cite{Bharti2025} that a notable sim-to-real transfer was demonstrated on a BlueROV  platform. TD3 was employed for visual servoing and docking, utilizing AprilTags for localization. Simultaneously, Zheng et al. (2025) \cite{Zheng2025} proposed a multi-layer simulation approach to bridge the reality gap. While their study targeted surface vehicles, their methodology—combining domain randomization with high-fidelity dynamics—is highly applicable to \acp{auv}, despite the remaining challenges of sensor noise and hardware failure.

The drive for efficiency led Chu et al. (2025) \cite{Chu2025} to develop an Adaptive Reward Shape PPO (ARSPPO) framework. By accounting for collision impacts and utilizing a parallel simulation framework to speed up training, they successfully demonstrated docking in a real-world lake environment. Further integration of computer vision was shown by Yu and Lin (2025) \cite{Yu2025}, who integrated a YOLO-based model for light-ring detection with a DDPG controller to perform docking in a test tank.

Most recently, Tuncay et al. (2026) \cite{Tunay2026-1, Tunay2026-2} have pushed the boundaries of training speed and validation. Their work highlights the necessity of realistic validation environments, where policies are trained in high-speed frameworks like MJX and subsequently validated in high-fidelity simulators like Stonefish. By utilizing the JAX-based MJX, they achieved 6-DoF control training in approximately five minutes using over 4,000 parallel environments. Their results indicated that SHAC and PPO could outperform standard MPC in controlled environments, marking a breakthrough in controlling complex \ac{auv} structures where standard mathematical models often struggle to predict behavior.

\section{Methodology}

    \subsection{Simulation}
    
\begin{figure}[htbp]
    \centering
    \begin{subfigure}{0.48\linewidth}
        \centering
        \includegraphics[width=\linewidth]{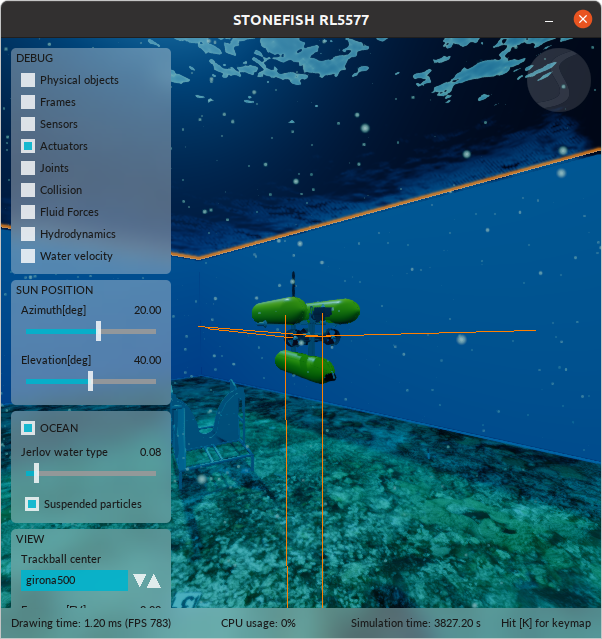}
        \caption{Digital-twin Environment}
        \label{fig:sim-env1}
    \end{subfigure}
    \hfill 
    \begin{subfigure}{0.48\linewidth}
        \centering
        \includegraphics[width=\linewidth]{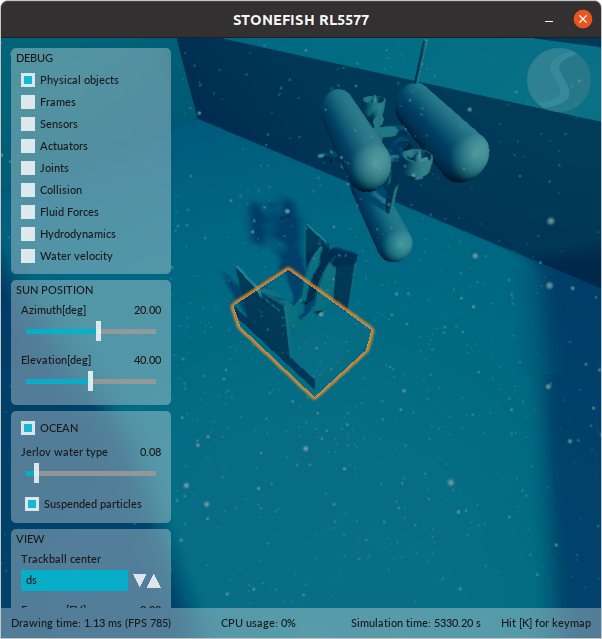}
        \caption{Physics object model}
        \label{fig:sim-collision}
    \end{subfigure}
    
    \caption{Overview of the simulation environment: (a) visual rendering and (b) collision geometry.}
    \label{fig:sim-total}
\end{figure}

To mitigate the sim-to-real gap during the transition from simulation to real-world deployment, we develop an accurate digital twin of the docking task using the Stonefish simulator \cite{stonefish, stonefish_ml,Font_stonefish_rl_2023}. \Cref{fig:sim-env1} presents the Girona AUV with the docking Station. Stonefish models the hydrodynamic behavior of the \ac{auv}, providing realistic vehicle dynamics. It also enables the direct integration of environmental disturbances such as currents, waves, and wind; however, in this work only ocean currents are considered as the primary perturbation. Additionally, the simulator provides realistic collision handling, allowing proper physical interaction and coupling between the \ac{auv} and the docking station during the docking maneuver, \cref{fig:sim-collision}. 

Stonefish-RL, allows to make simulations significantly faster than real-time. This can increase the speed of each thread up to 5 times. The variation in simulation speed depends on the CPU capabilities as it is the responsible for the physics computation (e.g: in case of collision, simulation slows down). Beside that, we implemented a multiprocess adaptation for Stonefish, which allows it to run in multiple-threads. For this Study we used 20 parallel threads plus one for evaluation. This values are less than ISAAC Sim or MJX which can reach up to 4096 instances but they ensure  realistic hydrodynamics and sensor models. The training threads runs headless (no graphical interface), while the evaluation use a graphical interface. This allows the trainer to observe the executed actions and the docking process. 

    \subsection{Docking Problem Setup}
    
In simulation, it is required to simplify the problem while keeping realistic assumptions of the docking problem. This is achieved in multiple phases. To learn the control behavior without being biased to a certain trajectory or path, the AUV and docking station spawning position is randomized in each episode. 

\begin{figure}
    \centering
    \includegraphics[width=0.8\linewidth]{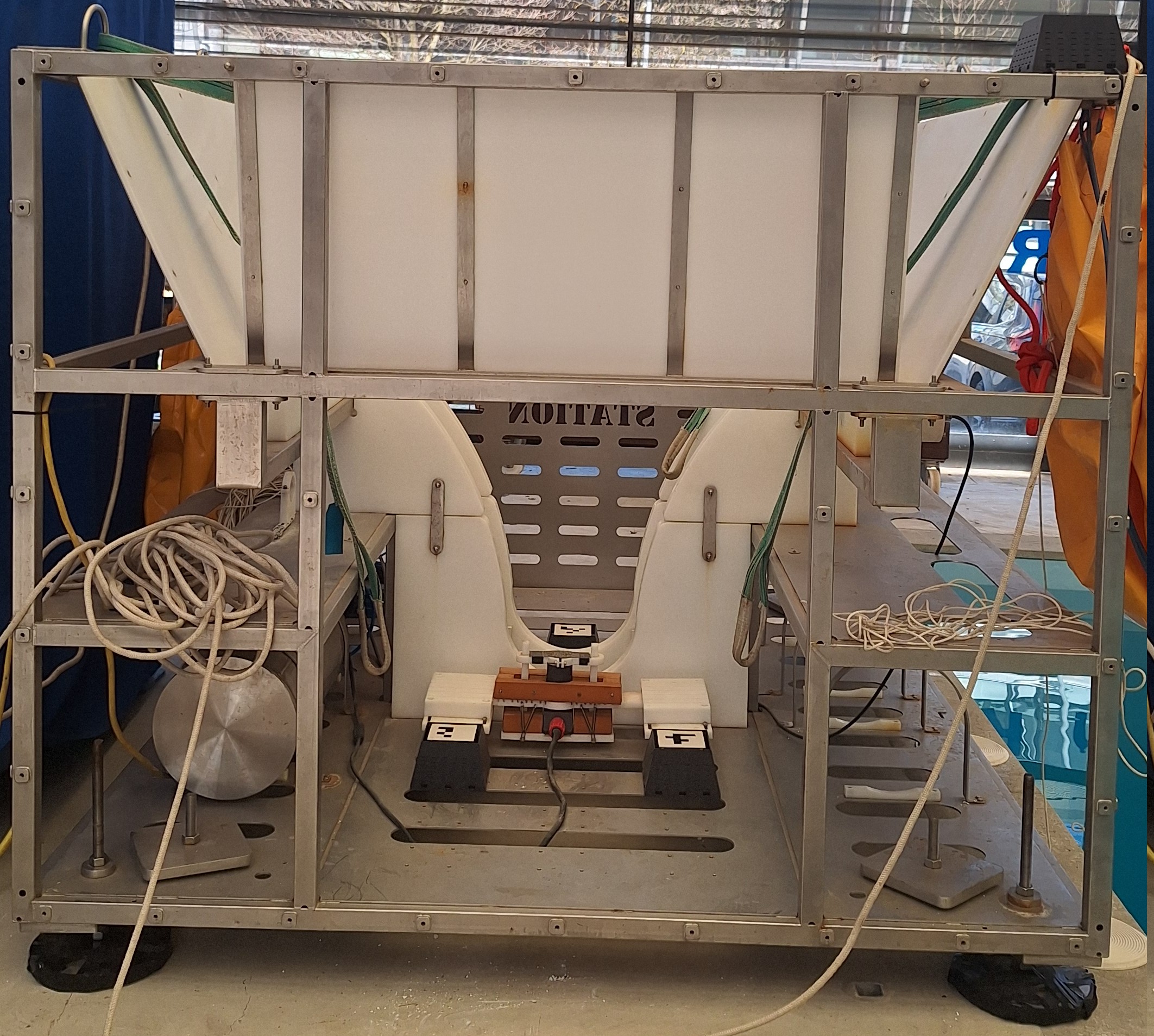}
    \caption{DS used in test tank}
    \label{fig:DS-img1}
\end{figure}

To ensure realistic collision handling and docking behavior, the \ac{ds} model includes all relevant collision points. As shown in \cref{fig:sim-total}, the simulation employs a simplified version of the \ac{ds} depicted in \cref{fig:DS-img1}. This simplified model includes only the guiding funnels, which facilitate docking by providing a clearance of $\pm 25$\,cm along the X and Y axes. The external metal frame that supports the guiding funnels, see \cite{chaarani2025docking}, is omitted in the simulation to simplify the meshes and improve computational performance.

To localize the \ac{ds}, position-based visual servoing is employed using an onboard camera and a \ac{3dbm} \cite{chaarani2025marker}. Since visual sensors are disabled during headless training (i.e., without a graphical interface), the camera model is simplified to a visibility condition. Specifically, when the \ac{ds} lies within the field of view of the \ac{auv}, its pose is updated with accurate measurements (see \cref{sec:state_space}). Prior to the first visual detection, we assume that an approximate estimate of the \ac{ds} pose, subject to uncertainty, is available through \ac{usbl} positioning or another external source.

The parameters used during training for the simulation, reward function, and model variables are presented in \Cref{tab:params} included in Appendix.\ref{appendix:params}.

\subsection{Algorithms And Policies}

The docking task is modeled as a \ac{mdp} defined by the tuple $(S,A,P,R,\gamma)$. At each time step k, the agent observes a state $s_k \in S$, selects an action $a_k \in A$ according to its policy $\pi$, receives a scalar reward $r_k=R(s_k,a_k)$, and transitions to the next state $s_{k+1}$ according to the transition probability $P(s_{k+1}|s_k,a_k)$
    
To learn the control policy $\pi$, we employ the Proximal Policy Optimization (PPO) \ac{drl} algorithm. PPO is an on-policy actor-critic method that employs a clipped surrogate objective to prevent large, destabilizing policy updates. PPO is selected for its reliability and ease of tuning in continuous control tasks. While Soft Actor-Critic (SAC) was also evaluated during the initial stages of this study, PPO demonstrated superior stability and performance during physical experimentation in the test tank. Consequently, PPO was selected as the primary algorithm for the final deployment presented in this work.

The objective of a \ac{drl} algorithm is to learn an optimal policy $\pi^*$ that maximizes the expected discounted return, defined as
\[
J(\pi) = \mathbb{E}\left[\sum_{k=0}^{\infty} \gamma^k r_{t+k}\right],
\]
where $\gamma \in [0,1)$ is the discount factor. The optimal policy is therefore given by $\pi^* = \arg\max_{\pi} J(\pi)$.

\subsubsection{State Space} \label{sec:state_space}

The state vector provided to the agent at each time step k is defined as:

\begin{equation} \label{equ:s_space}
    S = [\hat{\mathbf{o}}_k,e_\psi,V_k,A_k]
\end{equation}

where $\hat{\mathbf{o}}k$ represent the  translational error vector $[e_x,e_y,e_z]$ (relative position) of the docking point in the \ac{auv} body frame. To ensure a realistic state and facilitate sim-to-real transfer, Gaussian noise is injected into the observations based on distance and target visibility. The noise scale $\sigma_k$ is dynamically calculated relative to the Euclidean distance of the translational error:

\begin{equation}
\sigma_k = \frac{1}{6} | \mathbf{e}_{k, 1:3} |
\end{equation}

The perturbed observation $\hat{o}_k$ is then derived by adding two independent Gaussian components to the ground-truth observation $o_k$:
\begin{equation}
\hat{\mathbf{o}}_k = \mathbf{o}_k + \boldsymbol{\eta}{\text{base}} + \boldsymbol{\eta}{\text{occ}}
\end{equation}

where $\eta_{base} \sim \mathcal{N}\left(0, \left(\frac{\sigma_k}{2}\right)^2 I\right)$ represents constant sensor jitter. The occlusion noise $\eta_{occ}$ is conditioned on the visibility of the docking station $V_{ds} \in \{0,1\}$, such that $\eta_{occ} \sim \mathcal{N}(0, \sigma_k^2 I)$ if the target is not observed $V_{ds}=0$, and $\eta_{occ}$=0 otherwise. This approach prevents overfitting to perfect simulator coordinates by scaling uncertainty with distance and \ac{ds} visibility.

$e_\psi$ presents the yaw error in \ac{auv} frame. The variables $V_k$  presents the linear and angular velocities $v_x,v_y,v_z,\omega_\psi$ in the \ac{auv} frame. $A_k$ presents the \ac{auv} accelerations $acc_x,acc_y,acc_z$ measured by the \ac{imu}.

\subsubsection{Action Space}

The action space consists of a force and torque vector defined as
\begin{equation}
    A = [F_x, F_y, F_z, T_r, T_p, T_\psi].
\end{equation}
These forces and torques are expressed in the \ac{auv} body frame. The Girona \ac{auv} attempts to track the commanded wrench by distributing it among its five thrusters. Due to the vehicle's thruster configuration, the roll degree of freedom cannot be directly actuated. Nevertheless, a six-degree-of-freedom action vector is retained to maintain a general formulation.
    
\subsubsection{Reward Function}

    \begin{equation}\label{equ:reward}
        R = r_ {dist} + r_{angle} + r_{smooth}+r_{collision}+r_{mission}
    \end{equation}
    
\Cref{equ:reward} shows all the elements of the reward function. 
    \begin{equation} \label{equ:r_dist}
    r_{dist} = -\mathbf{w} \odot \mathbf{e} = -
    \begin{bmatrix} w_x \ w_y \ w_z \end{bmatrix} . 
    \begin{bmatrix} |e_x| \\ |e_y| \\ |e_z| \end{bmatrix}
    \end{equation}

$r_{dist}$ represents the Mahalanobis distance error between the \ac{auv} and \ac{ds} in the \ac{auv} frame. The distance is multiplied by the weight vector $[w_x,w_y,w_z]$, which allow the prioritization of certain axes as the docking requires. In this scenario, where a landing or vertical docking maneuver is considered, the X and Y axes are prioritized over the Z axis.

    \begin{equation} \label{equ:r_angle}
        r_{angle} =  \exp\left( -2 \cdot |E_\psi| \right) - 1 
    \end{equation} 

\begin{figure}
    \centering
    \includegraphics[width=\linewidth]{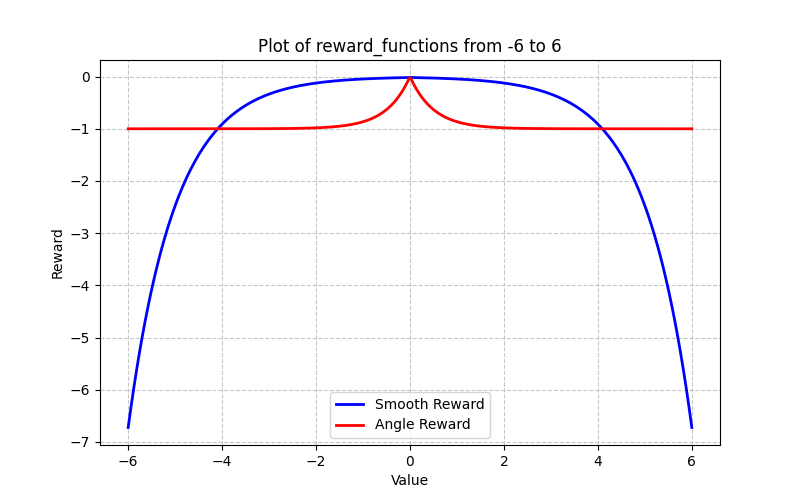}
    \caption{Exponential plots for the angle and smooth rewards}
    \label{fig:exp_plots}
\end{figure}

\Cref{equ:r_angle} penalizes the agent based on the yaw error, which is computed in the \ac{auv} body frame. For further illustration, \cref{fig:exp_plots} presents the behavior of this reward function.

\begin{equation}
r_{\text{smooth}} = -\frac{0.1}{N} \exp \left( \sum_{i=1}^{N} | a_{k,i} - a_{k-1,i} | \right)
\end{equation}

where $N$ represents the length of the action vector and $a_{k,i}$ denotes the $i$-th action component at time step $k$. This term penalizes large variations between consecutive actions, encouraging smoother transitions between control commands. \Cref{fig:exp_plots} illustrates the behavior of this reward function for variations between $-6$ and $6$, corresponding to the maximum possible change. This reward term also facilitates sim-to-real transfer, as smooth actuation is desirable for real-world deployment.

\begin{equation} \label{equ:r_smooth}
r_{\text{collision}} = 
\begin{cases} 
-p_c & \text{if } \|\mathbf{acc}_{\text{imu},k} - \mathbf{acc}_{\text{imu},k-1}\| > \Gamma_k \\
0 & \text{otherwise}
\end{cases}
\end{equation}
Collisions cannot be completely avoided during the docking process. Therefore, the objective is for the agent to enter the \ac{ds} softly while still using the guiding funnels. The parameter $p_c$ represents the collision penalty, as defined in \cref{equ:r_smooth}, and penalizes the agent based on impacts detected from acceleration variations. The parameter $\Gamma_k$ is an adaptive threshold defined as
\begin{equation}
\Gamma_{k+1} = 
\begin{cases} 
2\Gamma_k & \text{if a collision is detected} \\
\max(\Gamma_{set}, \Gamma_k / 2) & \text{no collision and } \Gamma_k > \Gamma_{set} \\
\Gamma_k & \text{otherwise}
\end{cases}
\end{equation} 

where $\Gamma_{set}$ represents the nominal threshold value. The adaptive threshold prevents sensor bouncing and avoids penalizing the agent multiple times for the same collision event. The decay step, implemented by dividing the threshold by 2, gradually returns it to its nominal value.

\begin{equation}
r_{\text{mission}} =
\begin{cases}
+p_s & \text{if } \text{goal is achieved} \\
-p_f & \text{if } \text{task is truncated} \\
0 & \text{otherwise}
\end{cases}
\end{equation}

The mission reward, $r_{mission}$, provides a high-magnitude terminal reward $p_s$ upon the \ac{auv} reaching the docking target within a defined proximity threshold, while applying a penalty $p_f$ for truncated episodes. This sparse reward is critical for reinforcing the global task objective, preventing the agent from settling for a local optimum—such as hovering near the target to maximize dense tracking rewards without actually completing the docking maneuver.

\section{results}

\subsection{Training}

\begin{figure}[htbp]
        \centering
        \includegraphics[height=4cm,width=\linewidth]{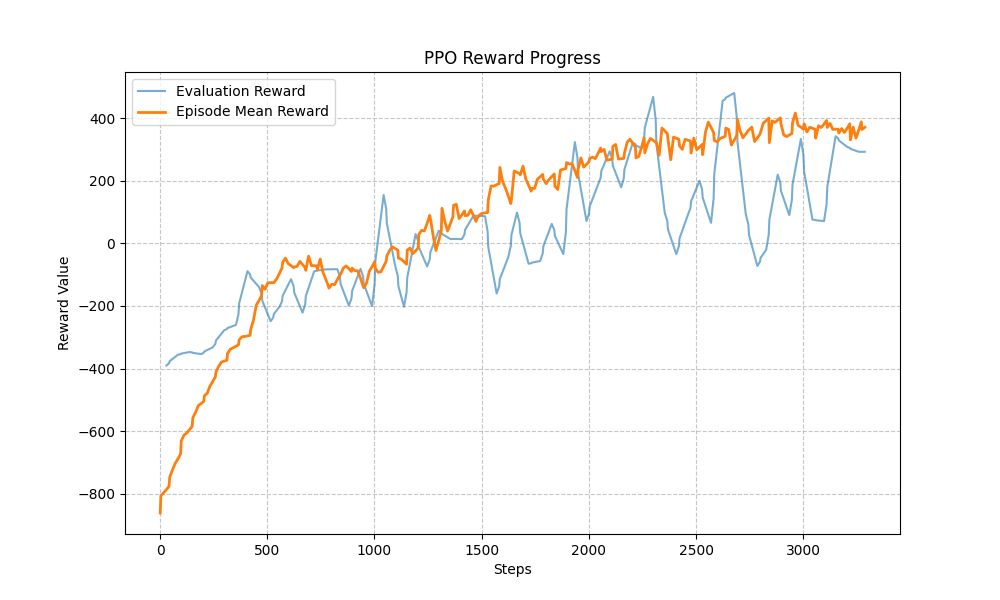}
        \caption{PPO algorithm}
    \caption{mean reward value during the training process using PPO agent}
    \label{fig:training}
\end{figure}

 \Cref{fig:training} presents the training process for the PPO agent. The training took around 3 hours using Intel Core i7 process with RTX 4060 Nvidia graphic card. The trained agent achieved over 90\% success docking rate with a mean reward between 300 to 400 depending on the spawn position while it was -800 at the start. 
 
\subsection{Simulation}
The simulations are performed using the Stonefish simulator integrated with ROS. Instead of interacting directly with the simulator, the learning agent communicates through the same ROS interfaces used by the real \ac{auv}. This design choice ensures that the software architecture in simulation closely matches that of the real system, minimizing the modifications required when transferring the policy to the physical vehicle. Although this ROS-based interaction introduces additional communication overhead and may slightly slow down the training process, it significantly facilitates sim-to-real transfer by preserving the same control and sensing interfaces used in real-world deployments.

\begin{figure}[htp]
    \centering
    \includegraphics[width=0.8\linewidth]{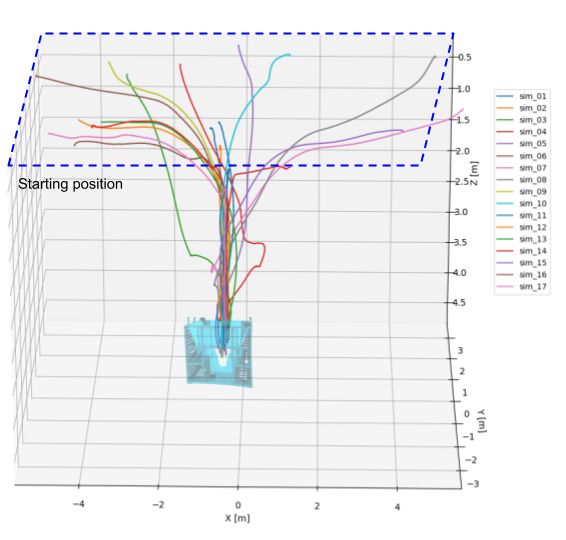}
    \caption{Simulated docking using RL Control}
    \label{fig:sim_traj}
\end{figure}

A downward-facing camera is used to estimate the position of the \ac{ds} by detecting a \ac{3dbm} \cite{chaarani2025marker}. The estimated pose is then transformed from the camera frame to the vehicle frame in order to compute the position error included in the observation vector.
\cref{fig:sim_traj} presents multiple docking maneuvers using the trained PPO agent. In these runs, the robot starting position is chosen randomly, causing the mission time to vary between 30 to 60 sec seconds. The DRL agent inference rate was set to 5\,Hz to match the camera processing rate as closely as possible.

\begin{figure}[htp]
    \centering
    \includegraphics[width=\linewidth]{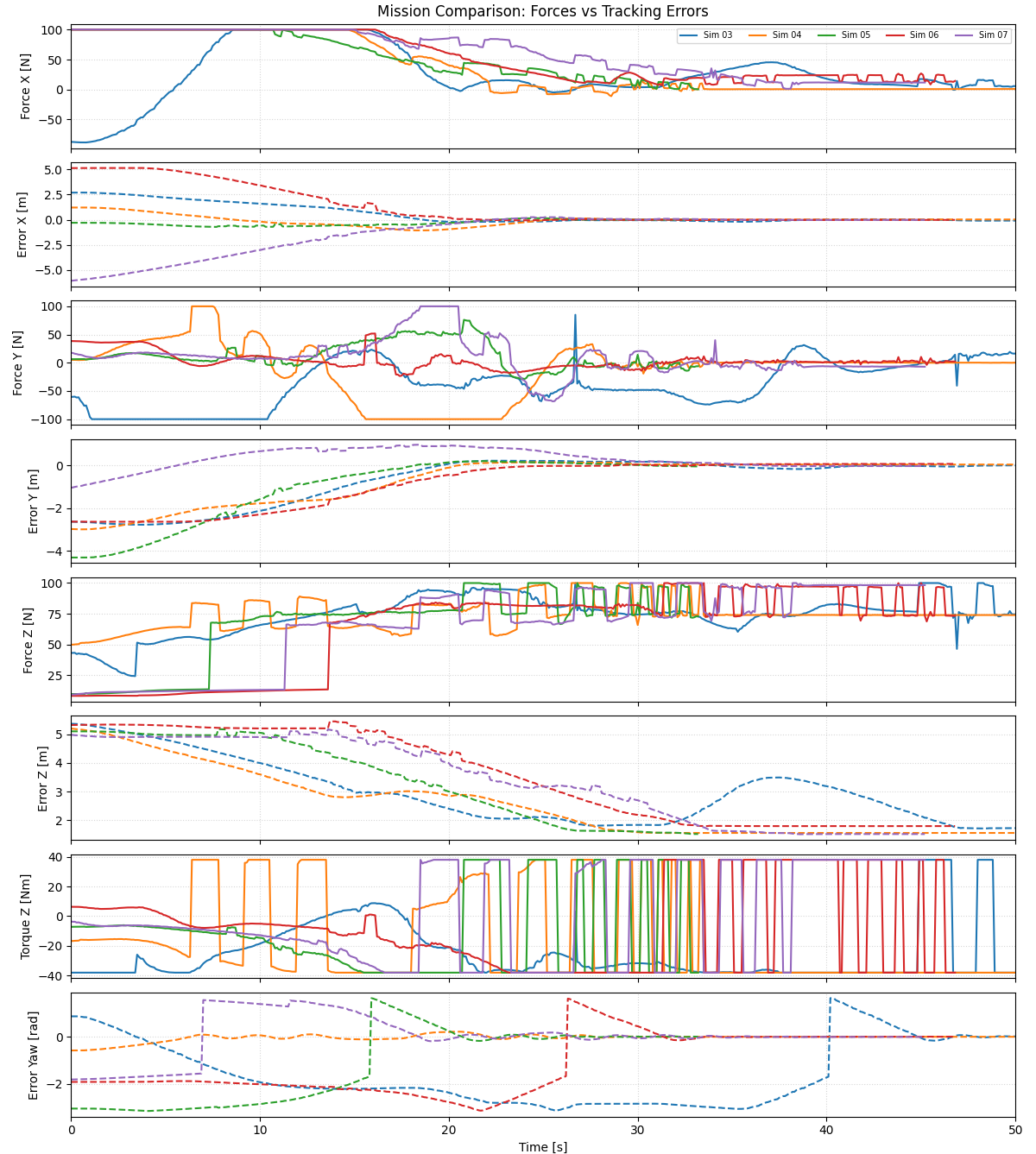}
    \caption{RL Force requests VS error in simulation}
    \label{fig:sim_forces}
\end{figure}

\Cref{fig:sim_forces} shows the forces and torques requested by the DRL agent together with the docking error in each axis. Five runs with a similar time range were selected for comparison. For easier visualization and interpretation, the error is plotted in the \ac{ned} frame. 
The errors in X and Y show a direct convergence with small overshoot. For the Z axis, the \ac{auv} begins to heave once the errors in X, Y, and $\psi$ have been significantly reduced. 
By inspecting the \ac{auv} behavior, we observed that two skills have been learned by the DRL agent that would be challenging to achieve with conventional controllers. The first is the use of pitch motion to brake and slow down. This behavior is observed during turning and diving as the \ac{auv} approaches the \ac{ds}.  
The second is the oscillation in yaw, observable in the torque plot in \cref{fig:sim_forces}. The authors conclude that this behavior helps the \ac{auv} slide inside the \ac{ds}. These small oscillations did not trigger the collision penalty, while still assisting the \ac{auv} in docking and entering properly.

\subsection{Test Tank Experiments}
\begin{figure}[hbtp]
    \centering
    \includegraphics[width=0.8\linewidth]{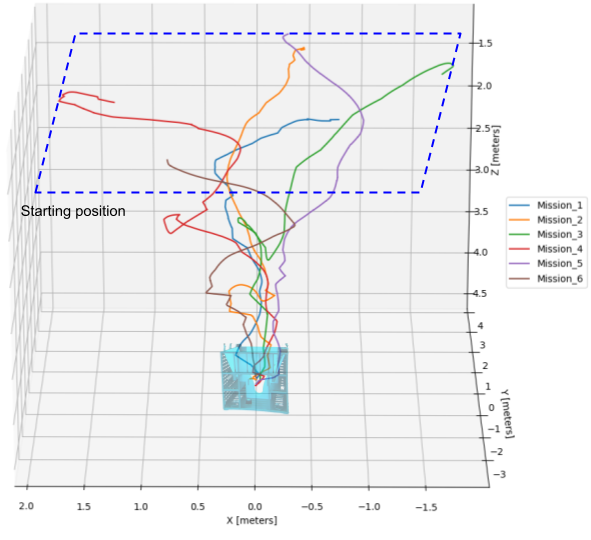}
    \caption{Test tank trajectories using RL control}
    \label{fig:real_traj}
\end{figure}

\Cref{fig:real_traj} displays the \ac{auv} trajectories obtained when executing the DRL agent on the real Girona \ac{auv} inside a test tank of size $19 \times 9 \times 5$~m. The experimental setup is similar to the simulated scenario. The \ac{ds} position is estimated using a downward-facing camera, after which the relative position is computed to compose the observation vector. Six different docking maneuvers were performed, each starting from different positions and random orientations. In total 10 missions are performed, eight were successful. \Cref{fig:real_traj} include six of these missions which took between 30 and 50 seconds. 

\begin{figure}[hbtp]
    \centering
    \includegraphics[width=1\linewidth]{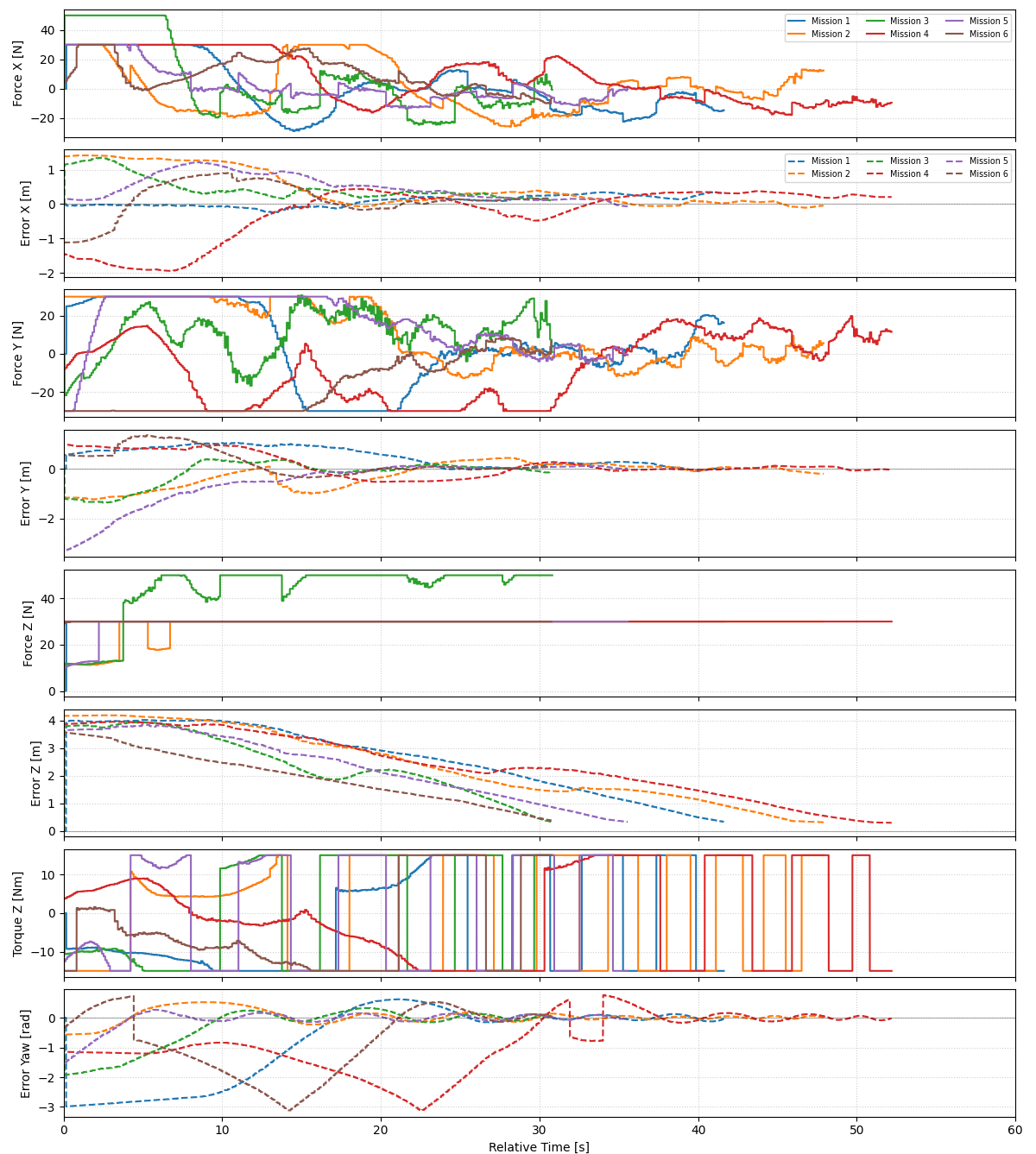}
    \caption{RL control force requests VS error in test tank }
    \label{fig:real_forces}
\end{figure}

\Cref{fig:real_forces} shows the forces and torques requested by the RL control agent together with the error in the \ac{ned} frame. For safety reasons, during the test tank experiments the forces were limited to $25\%$ or $50\%$ of the \ac{auv} maximum capabilities by clipping the values. 

\Cref{fig:real_forces} shows a behavior very similar to the simulated results, particularly the yaw oscillations observed during docking. This similarity suggests a successful sim-to-real adaptation.

\section{CONCLUSIONS}
This paper presented a comprehensive framework for transitioning Deep Reinforcement Learning policies from a high-fidelity digital twin to a physical \ac{auv} for docking tasks. By leveraging the Stonefish simulator in a multiprocessing environment, we successfully reduced training latency while maintaining the complex hydrodynamic characteristics of the Girona AUV.

Our results demonstrate that the DRL agent not only achieved a success rate exceeding $90\%$ in simulation but also exhibited a high degree of adaptability during physical test tank trials with 8 out of 10 successful runs. Crucially, the agent developed emergent tactical behaviors—such as using pitch angles for deceleration and yaw oscillations for mechanical alignment—without explicit programming. These behaviors proved vital in bridging the sim-to-real gap, allowing the AUV to navigate sensor noise and physical contact transitions that often destabilize traditional PID or MPC controllers.

Future work will focus on expanding the environmental complexity by introducing dynamic currents and dynamic docking stations. Also, we could randomize the thruster positioning during training to adapt to slight differences in the thruster setup between simulation and reality. Ultimately, this research confirms that high-fidelity simulation, when coupled with robust reward shaping, provides a reliable pipeline for deploying autonomous RL-based controllers in sensitive underwater environments.

\appendices

\section{Abbreviations}
\printacronyms[name= ]

\newpage

\section{Nomenclature and Acronyms}
\label{appendix:nomenclature}
\begin{table}[!htp]
\centering
\caption{Definition of Symbols used in Table \ref{tab:rl_SOA} }
\label{tab:acronyms}
\small
\begin{tabularx}{\columnwidth}{@{}l X@{}}
\toprule
\textbf{Symbol} & \textbf{Description} \\ 
\midrule 
\rowcolor[gray]{0.9} \multicolumn{2}{l}{\textbf{Algorithms and DRL Methods}} \\
PPO & Proximal Policy Optimization \\
SAC & Soft Actor-Critic \\
TD3 & Twin Delayed Deep Deterministic Policy Gradient \\
ABPPO & Adaptive Buffer PPO \\
ARDR & Adaptive Rollback Demo Replay PPO \\ 
ARSPPO & Adaptive Reward Shape PPO \\
DDPG & Deep Deterministic Policy Gradient \\
DQN & Deep Q-Network \\
DQNH/E & DQN with Hindsight Experience Replay / Extended \\
DroQ & Dropout Q-Functions for Doubly Robust Soft Actor-Critic \\
MPC & Model Predictive Control \\
SHAC & Short-Horizon Actor-Critic \\
YOLO & You Only Look Once (Object Detection Framework) \\

\midrule
 \rowcolor[gray]{0.9} \multicolumn{2}{l}{\textbf{State / Observation Variables}} \\
$x, y, z$ & Position coordinates in the global/local frame \\
$x_r, z_r$ & Relative position coordinates to the dock/target \\
$u, v, w$ & Surge, sway, and heave velocities (linear) \\
$\dot{x}, \dot{z}$ & Time derivatives of position (linear velocities) \\
$\theta, \psi, \phi$ & Pitch, yaw (heading), and roll angles \\
$\dot{\theta}, \dot{\psi}$ & Angular velocities \\
$r$ & Yaw rate (angular velocity about the Z-axis) \\
$\chi$ & Course angle (angle of velocity vector) \\
$d, d_t$ & Euclidean distance to the target at time $t$ \\
$c, k, c_d$ & Clearance, curvature, and drag/curvature coefficient \\
$\Delta d, \Delta \psi$ & Error in distance (x,y,z) and orientation (roll,pitch,yaw) \\
$e_x, e_y, e_z$ & Position errors in X, Y, and Z axes \\
$e_{y,t}$ & Cross-track error (lateral deviation from path) \\
$e_{1\times6}$ & 6-DoF pose error vector \\
$n$ & Propeller rotational speed (RPM) or discrete step \\
$T_u, T_v$ & Object pixel coordinates in camera \\
$W_b, h_b$ & Object pixel width \& height in camera\\


\midrule
\rowcolor[gray]{0.9}  \multicolumn{2}{l}{\textbf{Action Variables}} \\
$Q_m$ & Motor torque or propulsion command \\
$\delta_s, \delta_t$ & Control surface/rudder deflection angle \\
$n_1, n_2, n_3$ & Individual thruster speeds/commands \\
$f, f_t$ & Propulsive force (surge command) \\
$m_t$ & Control moment (torque) command \\
$F_x, F_y, F_z$ & Commanded forces in X, Y, and Z axes \\
$\tau_\phi, \tau_\theta, \tau_\psi$ & Commanded moments (Roll, Pitch, Yaw) \\

\midrule
 \rowcolor[gray]{0.9} \multicolumn{2}{l}{\textbf{Reward Components}} \\
$r_{dist}, r_{pos}$ & Reward/penalty based on distance to goal \\
$r_{align}, r_{att}$ & Reward for orientation/attitude alignment \\
$r_{thrust}, r_{act}$ & Penalty for excessive control effort/energy \\
$r_{smooth}$ & Penalty for non-smooth/jerky actuator movements \\
$e_{\delta H}$ & Penalty for depth/altitude error \\
$r_{act-mavg}$ & Reward based on action moving average smoothness \\
$r_{time}$ & Penalty for elapsed time (encourages efficiency) \\
\bottomrule
\end{tabularx}
\end{table} 

\vspace{2cm}
\section{Hyperparameters and Simulation Setup}
\label{appendix:params}

\begin{table}[!ht]
\centering
\caption{Hyperparameters for PPO and SAC training, and Simulation Environment Settings.}
\label{tab:params}
\begin{tabular}{lll} 
\hline
\textbf{Category} & \textbf{Parameter} & \textbf{Value} \\ \hline
\textit{Simulation} & Physics calculation freq. & 300 Hz \\
 & RL inference freq. & 5 Hz \\
 & Camera refresh rate & 5 Hz \\
 & AUV starting range & $\pm [6,3,1.4]$ \\
 & AUV starting Yaw range & $\pm \pi$ \\
 & DS starting range & $\pm [1,2,0]$ \\
 & DS starting Yaw range & $\pm \pi$ \\\hline
 \textit{reward} & $[w_x,w_y,w_z]$ & [1,1,0.5] \\
 & Collision penalty $p_c$ & -10 \\
 & Collision threshold $\Gamma_k$ & $1m/s^2$ \\
 & Successful docking $p_s$ & $+500$ \\
 & Failed docking $p_f$ & $-10$ \\ \hline
 
\textit{General RL} & Policy Network & MlpPolicy \\
& Max Episode Length & 60 s \\ 
 & Learning rate & $5 \times 10^{-4}$ \\
 & Buffer size & 500,000 \\
 & Batch size & 1024 \\
 & Gamma ($\gamma$) & 0.99 \\ \hline
\textit{PPO Specific} & Steps per Update ($n\_steps$) & 512 \\
 & Entropy Coeff ($ent\_coef$) & 0.01 \\
 & Clip range & 0.2 \\ \hline
\textit{SAC Specific} & Target smoothing ($\tau$) & 0.005 \\
 & Train frequency & 1 \\
 & Gradient steps & 1 \\
 & Learning starts & 1000 \\ \hline
\end{tabular}
\end{table}

\addtolength{\textheight}{-17cm}
\section*{ACKNOWLEDGMENT}

Alaaeddine Chaarani was supported by the Joan Oró Grant no. 2024 FI-1 00936. TANDEM research project funded by the MCIN/AEI/10.13039/501100011033 and the European Union. AI4AUV, Artificial Intelligence for AUV-based underwater habitat restoration research project (AIA2025-163346-C4) funded by the  Spanish Ministry of Science and Innovation.


\bibliography{bibtex}
\bibliographystyle{IEEEtran.bst}

\end{document}